\documentclass[preprint,12pt]{elsarticle}

\usepackage{hyperref}
\usepackage{tikz-dependency}
\usepackage{xcolor}
\usepackage{tikz}
\usepackage{listings}
\usepackage[ruled,linesnumbered,algo2e]{algorithm2e}
\usetikzlibrary{arrows,shapes,snakes,matrix,decorations.pathmorphing,backgrounds,fit,positioning,shapes.symbols,chains}
\definecolor{darkblue}{rgb}{0, 0, 0.5}
\hypersetup{colorlinks=true,citecolor=darkblue, linkcolor=darkblue, urlcolor=darkblue}

\begin{document}

\begin{frontmatter}

\title{Semantic Role Labeling for Knowledge Graph Extraction from Text}

\author[cnr-it]{Mehwish Alam}
\ead{mehwish.alam@istc.cnr.it}

\author[cnr-it,unibo]{Aldo Gangemi}
\ead{aldo.gangemi@istc.cnr.it}

\author[cnr-it]{Valentina Presutti}
\ead{valentina.presutti@istc.cnr.it}

\author[unica]{Diego Reforgiato Recupero}
\ead{diego.reforgiato@unica.it}

\address[cnr-it]{ISTC, CNR, Rome, Italy.}
\address[unibo]{University of Bologna, Bologna, Italy.}
\address[unica]{Universit\`a degli Studi di Cagliari, Cagliari, Italy.}


\begin{abstract}
This paper introduces {\tt TakeFive}, a new semantic role labeling method that transforms a text into a frame-oriented knowledge graph. It performs dependency parsing, identifies the words that evoke lexical frames, locates the roles and fillers for each frame, runs coercion techniques, and formalises the results as a knowledge graph. This formal representation complies with the frame semantics used in Framester, a factual-linguistic linked data resource. 
The obtained precision, recall and F1 values indicate that TakeFive is competitive with other existing methods such as SEMAFOR, Pikes, PathLSTM and FRED. We finally discuss how to combine TakeFive and FRED, obtaining higher values of precision, recall and F1.
\end{abstract}


\end{frontmatter}


\section{Introduction}

Most knowledge in linked data and knowledge graphs is of a relational nature: people participating in events, products having prices, artifacts with parts, works of art produced by artists, beers sold at a bar, etc. For that reason, a good part of integration and interoperability ends up consisting in aligning relations among heterogeneous schemas and data.

Less known is the fact that the relations holding between entities are usually part of a larger context or situation: beers can be found at a bar because there is a selling/purchase situation; artists produce works because there is a creative process involved; artifacts are assembled through craftmanship or industrial procedures, products are assigned prices in the market, people are assigned roles in events, etc.

Regardless of the representation language used  and its serialization, existing knowledge graphs share a common limit: they express facts that typically lack contextual and situational information. This limit makes interoperability difficult. When two different datasets need to be integrated, implicit situations need to be reconstructed: This happens quite smoothly in humans, but not in knowledge-based systems.

A method to contextualise knowledge graphs is to express the facts that they capture as \emph{projections of frames}. \emph{Frames} are cognitive structures that are used by humans for organising their knowledge, as well as for interpreting, processing or anticipating information (cf.~\cite{Gangemi2010} for a discussion encompassing both linguistic and knowledge-based approaches to frames). In linguistics, a reference model for frames is Fillmore's Frame Semantics~\cite{fillmore1976frame}, where a frame is introduced intuitively as ``a kind of outline figure with not necessarily all of the details filled in''. More precisely, a frame is a structure that reifies an n-ary relation with multi-varied arguments, denotes a situation, event, state, or configuration, and is supposed to bear representational similarity to the knowledge encoded in cognitive systems. Any binary projection of a frame is called a \textit{semantic role}. For example, in the  sentence \textit{I bought a pair of shoes}, the word ``bought'' identifies an occurrence of a commercial event, where ``I'' and ``pair of shoes'' are objects that play the roles of ``buyer'' and ```goods'' respectively in the Commerce\_buy frame.
Fillmore's Frame Semantics has been substantiated by FrameNet~\cite{Baker:1998}: a long-standing, manually developed resource of (English) frames represented in a structured format by a group of linguists in Berkeley.

Recently, two resources have been introduced which support semantic interoperability by using frames: FrameBase~\cite{rouces2015framebase} and Framester~\cite{GangemiAAPR16}. The idea is simple in principle: since situations are frame occurrences, let's align any schema to a set of frames from a stable ontology, and make data interoperate along that path (if a schema fragment 1 and a schema fragment 2 align to a same frame, the respective data can be jointly queried modulo ontology-based data access, where the ontology of frames is uniform across resources).
This is apparently good news, but while an initial ontology of frames can be found in FrameNet, the methods by which we can actually align any existing schema to frames is much less obvious. And the main reason is that the relations defined in schemas, ontologies, and knowledge graphs cannot be trivially aligned to semantic roles. 

For example, we need to assign the relation foaf:knows to a semantic role within frames such as framenet:Personal\_relationship or framenet:Familiarity. However, there is no semantic role corresponding to foaf:knows. The alignment would work only as a result of a path internal to the frame, e.g. an OWL property chain on roles, such as: $isPartnerIn~o~hasPartner$.

Current approaches are still struggling with this problem: FrameBase manually aligns relations to semantic roles, leading to  scalability issues, while Framester provides an extensive amount of linguistic mappings that help a semi-automated alignment, but a previous linguistic parsing of the relations and their context is required, which is still non-standard, specially considering that only a few ontologies explicitly encode the competency questions that led to the form of their relations.

In practice, the integration of a knowledge extraction approach from competency questions or other textual material, its alignment to Framester, and the usage of dereferencing methods as proposed by FrameBase, collectively seem a viable automated integration solution in the future.
In order to foster the research, in this paper we propose a knowledge-graph-based algorithm for labeling semantic roles from an arbitrary text, thus accommodating for the linguistic parsing needed to perform  frame alignment prior to interoperability. Our algorithm, called TakeFive, is evaluated, and compared to alternative approaches, with respect to metrics widely applied in two NLP tasks: \textit{frame detection} and \textit{semantic role labeling}. The first refers to the ability to automatically detect occurrences of frames in natural language text. The second refers to identifying the fragments of text denoting the entities that play specific roles in a frame occurrence.
In this paper, we extend our preliminary work~\cite{iswc2017diego}, and use Framester~\cite{GangemiAAPR16}, a frame-based knowledge graph, to address frame detection and SRL with TakeFive\footnote{\url{https://github.com/TakeFiveSRL/TakeFiveSRL}}. 
TakeFive uses NLP resources and software components, but integrates them in a semantic web pipeline that produces knowledge graphs ready to be used for interoperability across data and schemas.
We intend to verify if a knowledge-based hybrid method is comparable to purely statistical methods, while retaining the ability to extract a properly linked knowledge graph from a sentence.

\begin{figure}
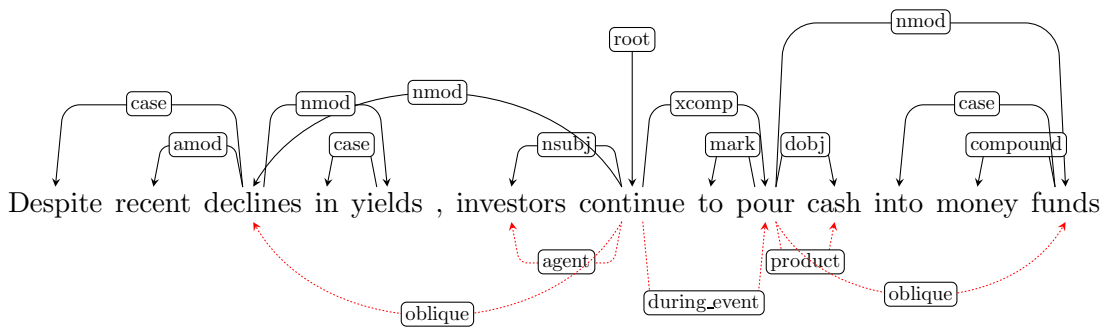

\small
\begin{center}
\begin{dependency}
\begin{deptext}
Despite \& recent \& declines \& in \& yields \& , \& investors \& continue  \& to  \& pour  \& cash  \&  into  \&  money  \&  funds \\
\end{deptext}
\deproot{8}{root}
\depedge{8}{7}{nsubj}
\depedge[edge below,red,densely dotted]{8}{7}{agent}
\depedge{8}{10}{xcomp}
\depedge[edge below,red,densely dotted]{8}{10}{during$\_$event}
\depedge[arc edge]{8}{3}{nmod}
\depedge[edge below,,arc edge,red,densely dotted]{8}{3}{oblique}
\depedge{3}{2}{amod}
\depedge{3}{1}{case}
\depedge{3}{5}{nmod}
\depedge{5}{4}{case}
\depedge{10}{9}{mark}
\depedge{10}{11}{dobj}
\depedge[edge below,red,densely dotted]{10}{11}{product}
\depedge{10}{14}{nmod}
\depedge[edge below,arc edge,red,densely dotted]{10}{14}{oblique}
\depedge{14}{13}{compound}
\depedge{14}{12}{case}
\end{dependency}
  \end{center}
  \vspace{-1cm}
  \caption{Dependency tree obtained from CoreNLP. Dotted lines represent the role labeling by TakeFive.}
      \label{fig:corenlp}
\end{figure}

As an example of the output of SRL, let us consider the following sentence 
\begin{equation}
\parbox{12cm}{{\textit{Despite recent declines in yields, investors continue to pour cash into money funds.}}}
\end{equation}
By performing frame detection, we recognize that \textit{to pour} evokes the frame \texttt{Pour.v} from Framester, subsumed by the frame \texttt{Cause\_motion} from FrameNet, meaning that the sentence expresses an occurrence of this frame. By performing SRL, TakeFive then labels \textit{the investors} and \textit{cash} respectively with the \texttt{Agent.cause\_motion} role, and the \texttt{Theme.cause\_motion} role, as both involved in the \texttt{Cause\_motion} situation occurrence (cf. Figure \ref{fig:corenlp}). The annotations use entities from reference ontologies for frames and semantic roles.

The remainder of the paper is organized as follows: Section~\ref{related} describes related work for SRL and SRL-based knowledge extraction, Section~\ref{resources} briefly introduces the data, resources and components used by TakeFive.  Section~\ref{method} presents the TakeFive algorithm. Section~\ref{results} describes the evaluation setting, the performance measures, and the results, with a comparison to other tools and a discussio.
\section{Related Methods}\label{related}

Semantic technologies usually leverage syntactic resources to improve their accuracy. Stanford CoreNLP\footnote{\url{https://stanfordnlp.github.io/CoreNLP/}}~\cite{manning-EtAl:2014:P14-5}, which we use in TakeFive, is one of the most used full-fledged NLP tools, and has been used in the Semantic Web context. However, it has not been extensively employed for SRL (an exception is \cite{kilgarrif2014}).

After the development of PropBank~\cite{Kingsbury2002}, where semantic information has been added to the Penn English Treebank data set, and the CoNLL shared tasks on semantic role labeling~\cite{carreras2004,Carreras2005}, there has been a lot of research in this domain, typically using PropBank as the reference ontology for roles. PropBank is a data set consisting of the phrase-structure syntactic trees from the WSJ section of the Penn Treebank. Its annotations include predicate-argument structures for verbs and defines a small number of roles: core roles are ARG0 through ARG5, which can be interpreted differently for different predicates. Further modifier roles ARGM* include e.g. ARGM-TMP (temporal) and ARGM-DIR (directional). 

The semantics of the core roles ARG0-ARG5 is not straightforwardly clear. 
The study described in ~\cite{Yi07cansemantic} shows that the roles ARG2-ARG5 serve many different purposes for different verbs, and points out that they are inconsistent and highly overloaded. In order to improve the performance for the SRL task, the arguments were mapped to VerbNet thematic roles. Others, e.g.~\cite{Bauer2011} revised the syntactic subcategorization patterns for FrameNet lexical units, using VerbNet. While PropBank labels the roles of verbs with a limited number of tags, frame-semantic parsing labels frame arguments with frame-specific roles, making it clearer what those arguments may mean. Therefore, for frame-semantic parsing, sentences may contain multiple frames that need to be detected along with their arguments. SemEval 2007 task 19~\cite{Baker2007} addressed this problem. The task leveraged FrameNet 1.3 and released a small corpus containing more than 2000 sentences with full text annotations.


The work described in~\cite{Furstenau2012} projects predicate-argument structures from seed examples to unlabeled sentences using linear program formulation to find the best alignment related to the projection. The projected information and the seeds are both used to train statistical models for SRL. In addition, the authors introduce a method for finding examples for unseen verbs using a graph alignment tool, which was used to project annotations from seed examples to unlabeled sentences.

In~\cite{Titov2012,Lang2011} the authors use an unsupervised approach for SRL that aims at inducing semantic roles automatically from unannotated data. Although this can be useful to discover new semantic frames and roles, in this paper we focus on the concrete representation provided in FrameNet and VerbNet, without expanding their inventory of semantic types.

Authors in~\cite{Giuglea2006} introduce a semantic parser that uses a broad knowledge base created by interconnecting FrameNet, VerbNet and PropBank. SEMAFOR~\cite{semafor}\footnote{\url{http://www.ark.cs.cmu.edu/SEMAFOR}} is a well known system for frame-semantic parsing, based on the combination of knowledge from FrameNet, two probabilistic models trained on full text annotations released along the FrameNet lexicon, and expedient heuristics. At SemEval 2007 it outperformed existing approaches.

FRED\footnote{\url{http://wit.istc.cnr.it/stlab-tools/fred}}~\cite{PresuttiDG12,fredswj} is the state-of-the-art tool for producing framed knowledge graphs for the Semantic Web. It consists of a complex pipeline of NLP and Semantic Web components for parsing text, representing it to a neo-Davidsonian logical form, extracting entities, disambiguating predicates, linking them to public resources, and creating a well-connected, formal and queryable knowledge graph out of that. FRED uses a ``greedy'' approach for SRL, i.e. it labels roles with reference labels (from either VerbNet or FrameNet) when the confidence of its categorial parser is high, otherwise it uses other heuristics to provide meaningful local labels that make sense in that textual context.

PIKES~\cite{pikes}\footnote{\url{http://pikes.fbk.eu}} is a tool that automatically extracts things of interest and facts about them from text. PIKES applies a number of NLP tools to annotate a text, and applies a linked-data-oriented approach to generate RDF graphs.

PathLSTM is a SRL system introduced in~\cite{RothL16}, which builds on top of the mate-tools semantic role labeler\footnote{\url{http://code.google.com/p/mate-tools/}}. It leverages neural sequence modeling techniques: the authors model semantic relationships between a predicate and its arguments by analyzing the  dependency path between a predicate word, and each argument head word. Lexicalised paths are considered, which are decomposed into sequences of individual items, namely  the  words  and  dependency  relations  on a  path. Long-short term  memory networks are then applied to find a recurrent composition function that can reconstruct an appropriate representation of the full path from its individual parts.

\section{Lexical resources used by TakeFive}\label{resources}
In this section we describe the lexical resources that we leverage in order to design TakeFive.

\subsection{VerbNet}\label{sec:verbnet}
VerbNet~\cite{Schuler:2005} is a broad coverage verb lexicon in English, with links to other data sources such as WordNet~\cite{fellbaum98wordnet} and FrameNet~\cite{Baker:1998}. It contains semantic roles and verb classes corresponding to Levin's classes \cite{levin1993english}, and including multiple verb senses. Verb classes can therefore be considered akin to word synsets. They generalise the verbs based on their shared syntactic behavior. These verb classes feature a simple two-layer hierarchy. For example, the verb \textit{conquer} is a member of the class \texttt{subjugate-42.3}, and hence a sense \texttt{Conquer\_42030000} is created (the sense of conquer in that class). 

VerbNet further contains semantic roles, which correspond to the relations between a verb sense and its arguments. Each class has multiple \emph{frames} (either syntactic- or semantic-oriented), which define a list of \emph{predicates} associated with their \emph{arguments}. There is a (partial) morphism between syntactic and semantic frames, so that semantic roles (``arguments'') are also associated with patterns that characterize the syntactic behavior of a verb in that class. For example, the roles defined for the class \textit{subjugate-42.3} are \emph{Agent}, \emph{Patient} and \emph{Instrument} meaning that an agent subjugates a patient with some instrument. Here \emph{Agent} and \emph{Patient} are necessary roles, and \emph{Instrument} is an optional role.
Verb senses help in determining if a particular verb instance conforms to the underlying semantics of the class. For the case of the verb \textit{conquer} its only sense is included in the class \texttt{subjugate-42.3}. VerbNet maps verbs to FrameNet frames, e.g. the verb sense \texttt{Conquer\_42030000} is mapped to the frame \texttt{Conquering}. The version of VerbNet used in TakeFive evaluation is 3.1, and the data come from the RDF porting of VerbNet 3.1 that is included in Framester\footnote{\url{http://github.com/framester/Framester}}~\cite{GangemiAAPR16}.

\subsection{FrameNet}\label{sec:framenet}
FrameNet~\cite{Baker:1998} contains \emph{frames}, which describe a situation, state or action. Each frame has semantic roles (``frame elements'') that are much more semantically detailed than VerbNet ones. FrameNet also defines a subsumption relation between either frames or roles. The subsumption relation can be used to create a hierarchy of classes, as shown in \cite{NuzzoleseGP11}. 
Each frame can be evoked by \emph{lexical Units (LUs)} belonging to different parts of speech. In version 1.5, FrameNet covers about 10,000 lexical units and 1024 frames. 
Let us consider the following sentence:
\begin{equation}\label{ex:running_example}
\parbox{12cm}{{\textit{[The Spaniards]$_{Conqueror}$ [conquered]$_{Lexical~Unit}$ [the Incas]$_{Theme}$.}}}
\end{equation}

In the above example, \textit{The Spaniards} is the argument (we will also refer to it as filler) of the role \textit{Conqueror} and \textit{the Incas} is the argument (or filler) of the role \textit{Theme} and \textit{conquered} is the lexical unit evoking the frame. 

\subsection{Framester}\label{sec:framester}
Framester~\cite{GangemiAAPR16} is a large RDF\footnote{\url{https://www.w3.org/TR/rdf11-primer/}} knowledge graph (currently including about 50 million  triples), acting as a hub between several predicate oriented linguistic resources such as FrameNet, WordNet, VerbNet \cite{Schuler:2005}, BabelNet \cite{Navigli:2012}, Predicate Matrix \cite{LacalleLR14}, as well as many other linguistic, factual, and foundational knowledge graphs. It leverages this wealth of links to create an interoperable and homogeneous \emph{predicate space} represented in a formal rendering of frame semantics \cite{fillmore1976frame} and semiotics \cite{Gangemi:10}. Framester uses a novel mapping between WordNet, BabelNet, VerbNet and FrameNet at its core, expands it to other linguistic resources transitively, and represents all of this formally. It further links these resources to other important ontological and linked data resources

Framester is accessible through its SPARQL endpoint\footnote{\url{http://etna.istc.cnr.it/framester2/sparql}}. Framester also features a subsumption hierarchy of semantic roles (i.e. frame elements) and adds generic roles on top of frame-specific roles. 

Framester also offers a Word Frame Disambiguation (WFD) service based on the mappings defined within the resource. It is available as a frame detection API which is implemented using UKB and Babelfy as the Word Sense Disambiguation algorithms as a first step and then the second step uses the mappings between WordNet/BabelNet synsets and FrameNet frames. The associated REST API is available online\footnote{\url{https://lipn.univ-paris13.fr/framester/en/wfd_html}}.
\section{Semantic Role Labeling Algorithm}\label{method}

TakeFive generates role oriented knowledge graphs given an input sentence. The algorithm begins by detecting the verb (lemma and VerbNet verb class), along with its arguments, and then it relates it to their corresponding VerbNet roles. According to our running example~\ref{ex:running_example}, TakeFive detects the verb \textit{conquered} and then extracts the VerbNet roles of this verb i.e., \textit{Conqueror} and \textit{Theme}. Finally, it assigns the role fillers i.e., \textit{The Spaniards} as a filler of \textit{Conqueror} and \textit{the Incas} as the filler of the VerbNet role \textit{Theme}. 
The backbone of TakeFive follows a step-wise approach:
{\bf (i)} preprocessing step for extracting dependencies and frame annotations using existing tools (i.e., CoreNLP and Word Frame Disambiguation respectively), {\bf (ii)} detecting (CoreNLP-derived, mainly syntactic) interface roles, {\bf (iii)} VerbNet specific roles (mainly semantic) for a certain frame, and then finally, {\bf (iv)} checking the compatibility between interface and semantically specific roles. In other words, we aim at using background knowledge and formal reasoning in order to associate semantic roles with syntactic dependencies. In the rest of the paper, we use the following terminologies:
\begin{itemize}
\item[--] {\textit{CoreNLP interface roles} for the roles generalising CoreNLP dependencies as well as resource-specific semantic roles;}
\item[--] {\textit{VerbNet specific roles} for the VerbNet roles related to a certain verb sense;}
\item[--] {\textit{VerbNet interface roles} for the roles that subsume VerbNet specific roles in Framester, and are subsumed by an interface role.}
\end{itemize}

\subsection{Pre-processing step}


\paragraph{Framester and CoreNLP} For a given input sentence, frame detection using Word Frame Disambiguation (WFD) is performed. It uses Babelfy as a WSD algorithm and then uses the mappings between BabelNet Synsets and FrameNet frames as given in Framester. 
The dependency tree is extracted associated to a given input sentence using CoreNLP. 
Listing~\ref{tripleset2} shows a dependency tree returned by CoreNLP for the running example.

\begin{lstlisting}[float,caption=Dependency tree triples representation of CoreNLP,label=tripleset2]
det, Spaniards-2, The-1,
nsubj, conquered-3, Spaniards-2, 
root, ROOT-0, conquered-3,
det, Incas-5, the-4,
dobj, conquered-3, Incas-5
\end{lstlisting}

%
%
%
%
%

\paragraph{Assigning Interface roles to CoreNLP dependencies} TakeFive is based on 23 simple heuristics for mapping CoreNLP dependency triples to interface roles. For the running example, we have a dependency {\tt nsubj, conquered-3, Spaniards-2} related to the verb {\tt conquered}, and its argument {\tt Spaniards}.  Dependency types such as {\tt nsubj, dobj,...} are generalized to \emph{CoreNLP interface roles} through a set of heuristics e.g., by applying the rule $nsubj \rightarrow Agent$ i.e., the role {\tt Agent} is assigned to the argument \textit{Spaniards}. The set of interface roles include \textit{\{Agent, Undergoer, Recipient, Eventuality, Oblique\}}. 


\subsection{TakeFive: A Semantic Role Labeling Algorithm}

This section discusses the two algorithms proposed for labeling a given sentence with the VerbNet specific and VerbNet interface as well as a way to check the the compatibility between the CoreNLP interface roles (as assigned previously) and VerbNet interface roles. Algorithm~\ref{alg1} computes VerbNet interface and specific roles of extracted verbs from an input sentence.

\subsubsection{Computing VN Interface and Specific Roles}

Algorithm \ref{alg1} takes the pre-processed information as an input i.e., (a) the sentence, (b) dependency tree obtained by CoreNLP and (c) the output of frame annotations obtained using WFD. It then returns the input sentence labeled with VerbNet specific as well as VerbNet Interface roles. If the verb is polysemic then it uses frame detection for extracting VerbNet roles (line 2-4, see Algorithm \ref{alg2}), otherwise it gets the verb sense using the SPARQL query in listing \ref{lst:Q1bis}. If it returns more than one verb sense it selects the one which is most frequent (see query in listing~\ref{lst:Q8}) and extracts the VerbNet specific roles along with VN interface roles, if any (line 6-9, see listing \ref{lst:Q4}). If the result is empty, it uses frame detection for obtaining the VerbNet roles described in Algorithm~\ref{alg2}.

Algorithm~\ref{alg2} and \ref{alg3} are used in case of polysemous verbs. The algorithm takes as an input a sentence and annotates it with frames (line 1). If there are no frame annotations it takes the most frequent verb sense using SPARQL query in listing \ref{lst:Q8} and then the VN specific and interface roles associated to the this verb sense through SPARQL query in listing \ref{lst:Q4} (line 2-5). If the Word Frame Disambiguation API returns multiple frames and there is a relation between these frames, the most specific frame is chosen (line 7-16) (see Algorithm~\ref{alg3}). Then, given the $verb$ and the chosen frame, VerbNet senses are extracted using listing~\ref{lst:Q3} (line 17). If no verb senses are returned then the most frequent verb senses are extracted for getting the VerbNet specific roles (line 18-20). However, if there are more than one verb senses then the most frequent verb sense is chosen. An intersection of both the queried and returned sets of verb senses is taken and the verb sense with highest ranking based on frequency of verb senses in WordNet is selected and the corresponding VN role is returned (line 22-30). If both the above cases are false the VN role associated to the verb sense is selected (line 32).

   \begin{algorithm2e}[htp!]
  \KwData{CoreNLP and Framester pre-processed information of an input sentence; an input sentence}
  \KwResult{Pairs (VerbNet interface roles, VerbNet specific roles) for each verb senses of the extracted verbs in the input sentence}
 \ForEach{verb} {

   \If{verb {\bf is} polysemic}{
   see Algorithm~\ref{alg2}
    }
   \Else{
   $verbSenses \leftarrow $ get verb sense using query in listing \ref{lst:Q1bis}\;
   \If{ $verbSenses != \emptyset$}{
       		$verbSense \leftarrow $ obtain most frequent verb senses using SPARQL query in listing  \ref{lst:Q8} \; 
		$vnrole \leftarrow$ given verbSense extract VN roles using SPARQL query in listing \ref{lst:Q4} \; %
   } 
   \Else{
   see Algorithm~\ref{alg2}
   }
   }
 }
   \caption{Computation of pairs (VerbNet interface roles, VerbNet specific roles) for each verb in a given sentence.}
  \label{alg1}
 \end{algorithm2e}

   \begin{algorithm2e}[htp!]
  \KwIn{sentence, verb}
  \KwOut{3-tuple containing verb, its VerbNet specific role, VerbNet Interface role}
   	$frames  \leftarrow getFrames(sentence)$\;
   	\If{ frames {\bf is} $\emptyset$}{
           $verbSense \leftarrow $ obtain most frequent verb senses using query \ref{lst:Q8} \; 
           $vnrole \leftarrow$ given verbSense extract VN roles using query \ref{lst:Q4} \; %
    	}
        
   \caption{Obtaining VerbNet Roles using Frame Detection if no frames are obtained.}
  \label{alg2}
 \end{algorithm2e}       

\begin{algorithm2e}[htp!]
\setcounter{AlgoLine}{5}
    		\For{$f_1 \in frames$}{
     		\For{$f_2 \in frames$}{
       			\If{ $f_1 \sqsubseteq f_2$ }
        				{$frameSpecific \leftarrow f_1$} 
                  \ElseIf{ $f_2 \sqsubseteq f_1$ }
        				{$frameSpecific \leftarrow f_2$}
      		}
    		}
		$verbSenses \leftarrow $ given $verb$ and  $frameSpecific$ using query \ref{lst:Q3}\;
		\If{ $verbSenses == \emptyset$ } 
    	        { $verbSense \leftarrow $ obtain most frequent verb senses using query \ref{lst:Q8} \;
		$vnrole \leftarrow$ given verbSense extract VN roles using query \ref{lst:Q4} }
		\ElseIf{ $len(verbSenses) > 1$ }{
		$verbSenseQueried \leftarrow $ obtain most frequent verb senses using query \ref{lst:Q8} \;
			\If{ $verbSenseQueried \in verbSenses$ } 
			{$vnrole \leftarrow$ given verbSense extract VN roles using query \ref{lst:Q4} }
			\Else{
			$verbSense \leftarrow $ get the next most frequent verb sense obtained by query \ref{lst:Q8} and so on until the verbSense matches. 
			
			}
		}
		
                \Else{ 
		$vnrole \leftarrow$ given verbSense extract VN roles using query \ref{lst:Q4} 
                }
 
  \caption{Obtaining VerbNet Roles using Frame Detection if frames are detected.}
  \label{alg3}
 \end{algorithm2e}



\subsubsection{Checking Compatibility of CoreNLP interface roles}


The objective here is to return, all roles and fillers for each argument of verbs from the input sentence if the interface roles assigned using the two methods (i.e., heuristics and algorithm~\ref{alg1}) are compatible. Let $O = \{Agent,~Undergoer,$ $Recipient,~Eventuality\}$, $C$ be the CoreNLP interface roles (assigned using heuristics such as  $nsubj \rightarrow Agent$), $V$ be the VerbNet interface roles and $R$ be the VerbNet specific roles; where $V$ and $R$ are returned by algorithm~\ref{alg1}  for a given sentence.  For $v_1 \in V$, $c_1 \in C$ and $r_1 \in R$, if $v_1 = \emptyset$ and $r_1 = \emptyset$, then $c_1$ is assigned. However, if  $v_1 \neq \emptyset$ and $r_1 \neq \emptyset$, then the following rules are defined:

\begin{itemize}
\item[--] The algorithm starts by choosing $verb$ having at least one VN sense and takes $c_1$ and $v_1$. If $c_1\in O$ and $v_1 \in O$ then the pair $(c_1,v_1)$ is marked compatible and $r_1$ is returned such that $r_1 \in R$ and $v_1$ {\tt associated\_to} $r_1$ (returned by Algorithm~\ref{alg1}).

\item[--] If $c_1$ and $v_1$ associated with $verb$ are \textit{oblique} check for a preposition along with CoreNLP dependencies triples having the modifier \emph{nmod} and then return $r_1$ such that $r_1$ is compatible with the preposition according to VerbNet verb arguments. The association between the VerbNet arguments and the prepositions are already defined in VerbNet and now standardized in RDF in Framester linguistic linked data hub as shown in Listing~\ref{tripleset3} (we also make use of the Preposition Project dataset, which is another linked dataset in Framester). 

\item[--] If $c_1 = Agent$ or  $c_1 = Undergoer$ and $c_1 \neq v_1$ then select the top role of the  subsumption hierarchy associated to VerbNet interface role (defined by the predicate {\tt fschema\footnote{\url{https://w3id.org/framester/schema/}}:subsumedUnder}). If the top role is \emph{Theme} then select $v_1$.

\item[--] If all the  previously defined rules are false or if there is no mapping between $c_1$ and $v_1$ then return $c_1$.

\end{itemize}

\lstset{basicstyle=\small}
\begin{lstlisting}[basicstyle=\small,float,caption=Triples for obtaining VerbNet arguments.,label=tripleset3,frame=single]
vsprepositionselection:Conquer_42030000-with-Instrument
        a  vnschema:SensePrepSelection ;
        vnschema:hasGenericArgument  vndata:Instrument ;
        vnschema:hasPreposition	vndataprep:with ;
        vnschema:hasVerbSense	vndata:Conquer_42030000 .
\end{lstlisting}

 \begin{lstlisting}[basicstyle=\small,caption={Query for extracting Verb Senses given a verb lemma.},label={lst:Q1bis},frame=single]
 SELECT DISTINCT ?verbsense 
 WHERE { 
 	?verbsense rdfs:label ?verblemma ; 
        	   a fschema:VerbSense .
	filter regex(?verblemma, "conquer", "i") }
 \end{lstlisting}



\begin{lstlisting}[basicstyle=\small,caption={Query for mapping frames to verb senses when the verb is polysemous in VerbNet.},label={lst:Q3},frame=single]
SELECT DISTINCT ?verbsense
WHERE {
	?verbsense rdfs:label ?verblemma ; 
	             a vn31schema:VerbSense; 
	             skos:closeMatch ?frame }
 \end{lstlisting}
 
 
\begin{lstlisting}[basicstyle=\small,caption={Query to retrieve the VerbNet role for a certain (optional) interface role and a certain verb sense.},label={lst:Q4},frame=single]
SELECT DISTINCT ?interfacerole ?verbnetrole
WHERE {
   ?verbnetrole a vn31schema:Argument; 
   vn31schema:inVerbSense <VERBNET SENSE>
   OPTIONAL 
   { 
    ?verbnetrole fschema:subsumedUnder+ ?interfacerole .
	?interfacerole a fschema:InterfaceRole } }
 \end{lstlisting}

\begin{lstlisting}[basicstyle=\small,caption={Query to retrieve most frequent verb senses given a verb lemma.},label={lst:Q8},frame=single]
SELECT DISTINCT ?verbsense
WHERE {
	?verbsense rdfs:label <verblemma> ; 
		a vn31schema:VerbSense .
	?verbsense skos:closeMatch ?wnsense .
	?wnsense schema:tagCount ?freq .
	FILTER ((datatype(?freq)) = xsd:int) }
	ORDER BY DESC(?freq) LIMIT 1
 \end{lstlisting}

\section{Evaluation}\label{results}
This section details the experimental setting, and the two evaluation procedures for measuring the performance of the TakeFive SRL algorithm. It also describes a comparison between TakeFive and other SRL tools.

\subsection{Implementation details}\label{implementation}
The algorithm is developed in Python and uses REST-APIs for Framester and Stanford CoreNLP. It also employs Py4J\footnote{\url{https://www.py4j.org/}} as a bridge between Python and Java. 
A Java class was developed which can directly be called from the main Python code through Py4J. It can be faster if a cache mechanism is used to store Framester results, and SPARQL queries to the Framester endpoint. The TakeFive SRL tool is available on-line\footnote{\url{https://github.com/TakeFiveSRL/TakeFiveSRL}}. 

\subsection{Evaluation setting}
The performance evaluation was conducted for verifying if the chosen VerbNet roles associated with fillers are correct or not. We used the WSJ section of the Penn Treebank annotated with VerbNet and PropBank labels\footnote{\url{https://github.com/ibeltagy/pl-semantics/blob/master/resources/semlink-1.2.2c/1.2.2c.okay}}. These annotations include the VerbNet and PropBank roles associated with each verb of each sentence of the dataset, and related to each filler. As an example, consider the following sentence contained in the WSJ annotated dataset:

\begin{equation}\label{ex:running_example2}
\parbox{12cm}{{\textit{The Canadian pig herd totaled 10,674,500 at Oct. 1, down 3 from a year earlier, said Statistics Canada, a federal agency.}}}
\end{equation}

The two verbs \textit{totaled} and \textit{said} are indicated in the annotations together with their VerbNet verb classes, as well as their VerbNet and PropBank roles and fillers. In particular, Table~\ref{annotations2} shows  the annotations for sentence~\ref{ex:running_example2}.

 \begin{table}[]
 \centering
  \resizebox{1\textwidth}{!}{
 \begin{tabular}{|l|l|l|l|l|}
 \hline
\textbf{Verb} & \textbf{Verb class} & \textbf{VerbNet Role} & \textbf{PropBank Role} & \textbf{Filler}  \\ \hline
Say & 37.7-1 & Topic & ARG1 & The Canadian pig herd totaled \\ 
&  &  &  & 10,674,500 at Oct. 1, down 3\\
&  &  &  & from a year earlier \\ \hline
Say & 37.7-1 & Agent & ARG0 & Statistics Canada, a federal agency \\ \hline
Total & 54.1-1 & Theme & ARG1 & The Canadian pig herd \\ \hline
Total & 54.1-1 & Value & ARG2 & 10,674,500 \\ \hline
\end{tabular}}
 \small
 \caption{Annotations for example sentence~\ref{ex:running_example2}}
 \label{annotations2}
\end{table}

Performance evaluation was conducted by computing precision, recall and F1 score using the official CoNLL-2009 scorer\footnote{\url{https://ufal.mff.cuni.cz/conll2009-st/scorer.html}}~\cite{conll2009}. The CoNLL-2009 scorer evaluates the semantic frames by reducing them to semantic dependencies. A semantic dependency from every predicate to all its arguments is created. These dependencies are labeled according to their corresponding arguments. Additionally, a semantic dependency from each predicate to a virtual ROOT node is added. The latter dependencies are labeled with the predicate senses. This approach guarantees that the semantic dependency structure forms a single-rooted, connected (but not necessarily acyclic) graph. It can be seen that the scoring strategy gains some points even though a system assigns the incorrect predicate sense. For further details refer to~\cite{conll2008,conll2009}. 

In order to use the CoNLL scorer when comparing to other methods, we formatted the output of TakeFive as well as that of the other tools as required by the CoNLL-2009 scorer. Finally, for compliance purposes, we employed SemLink\footnote{\url{https://verbs.colorado.edu/semlink/}} to map VerbNet roles to PropBank roles.

\subsection{Results}
The results obtained by TakeFive have been compared to other state-of-the-art methods, including SEMAFOR, Pikes, FRED, and PathLSTM (for details (see Section~\ref{related})). 
As FRED and TakeFive are two resources maintained by overlapping teams, we have also combined their results by including all the VerbNet roles extracted by FRED into  the results of TakeFive; we named this new algorithm as TakeFive+FRED.
First of all, the \textit{onf} files (2454) of the WSJ corpus\footnote{The dataset was made available by the Linguistic Data Consortium: \url{https://www.ldc.upenn.edu}.} were processed. Each file contains input sentences and their parse trees. The gold standard has 74977 rows. Each row corresponds to a verb in a given sentence and includes the VerbNet verb class, fillers and VerbNet and PropBank roles associated to that verb. Different rows might refer to the same sentence (as there might be several verbs within a given sentence). As already mentioned, when labeling each sentence using TakeFive, we extracted Framester and CoreNLP information (frames, dependency triples, POS tags, etc.). To speed up the experiments, we used a cache mechanism so that the information is downloaded only once (the cache mechanism is not currently available in the on-line version of TakeFive software). 

Table~\ref{tabres1} shows the labeled and unlabeled precision, recall and F1-measure values of TakeFive, TakeFive+FRED, and the competitors, SEMAFOR, Pikes, PathLSTM and FRED. \emph{Labeled scores} are related to the correct identification of labeled dependency whereas \emph{Unlabeled scores} do not take into account labels. For example, for the correct proposition: \textit{verb.01: ARG0, ARG1, ARGM-TMP}, the system that generates the following output for the same argument tokens \textit{verb.02: ARG0, ARG1, ARGM-LOC} receives a labeled precision score of 2/4 because two out of four semantic dependencies are incorrect: the ROOT dependency is labeled "02" instead of "01" and the dependency to the "ARGM-TMP" is incorrectly labeled "ARGM-LOC". On the other hand, the same example would receive an unlabeled precision score of 4/4. SEMAFOR is the method with the highest accuracy. Our proposed approach is the second best, however the numbers are very close to that of SEMAFOR. We noticed that FRED has some internal issues with the offset extraction of words of the sentence and this affects the output representation. As far as Pikes is concerned we have observed that it misses some important roles for tokens. Its SRL engine is based on mate-tools\footnote{\url{http://code.google.com/p/mate-tools/}}, further developed in~\cite{Bjorkelund:2009:MSR:1596409.1596416} in 2009.
Similar conclusions, and similar performances with Pikes, can be drafted for PathLSTM which is also based on mate-tools.

The combined approach TakeFive+FRED is able to slightly outperform SEMAFOR. FRED captures complementary roles which TakeFive is unable to detect. Based on this intuition, we believe that future experiments on optimal combination of multiple SRL approaches might yield the best results.



\begin{table}[]
 \centering
   \resizebox{1\textwidth}{!}{
 \begin{tabular}{|l|l|l|l|l|l|l|}
 \hline
\textbf{Method} & \textbf{Lab. } & \textbf{Lab.} & \textbf{Lab. } & \textbf{Unlab.} & \textbf{Unlab. } & \textbf{Unlab.}   \\ 

\textbf{} & \textbf{Prec.} & \textbf{Recall} & \textbf{F1} & \textbf{Prec.} & \textbf{Recall} & \textbf{F1}   \\ \hline


 TakeFive &  80.12\% & 76.04\% & 78.02\% &  85.09\% & 80.44\% & 82.70\%\\ \hline
 SEMAFOR  & 81.05\% & 77.01\% & 78.97\% & 87.32\% & 82.97\% & 85,09\% \\ \hline
  TakeFive+FRED & \textbf{82.55\%} & \textbf{78.48\%} & \textbf{80.46\%} &  \textbf{87.60\%} & \textbf{83.18\%} & 
  \textbf{85.33\%}\\ \hline
 FRED & 74.02\% & 72.36\% & 73.18\% & 83.11\% & 81.65\% & 82.37\%\\ \hline
 Pikes & 72.11\% & 70.62\% & 71.35\% &  79.27\% & 78.15\% & 78.70\%\\ \hline
PathLSTM &  73.66\% & 71.65\% & 72.64\% & 82.64\% & 80.63\% & 81.62\%\\ \hline
\end{tabular}}
\small
 \caption{Labeled and unlabeled precision, recall and F1 values of TakeFive, TakeFive+FRED, SEMAFOR, Pikes, FRED and PathLSTM.}
 \label{tabres1}
 \end{table}

 \paragraph{A different strategy?} While the results with CoNLL indicate that TakeFive performs as good as state-of-the-art methods, and better in ensemble, in a semantic web context the evaluation strategy may be too lightweight. In order to test this, a different evaluation strategy has been conducted (not detailed here for space reasons, see \cite{iswc2017diego}), which follows more closely the kind of SRL extraction that is supposed to be represented in a knowledge graph. The results with this second evaluation strategy show a lower accuracy (more than half than the one obtained with the CoNLL scorer) because i) in strategy 2 we defined the score so that a matching is verified only when the role filler contains all the exact words of the gold standard, ii) in strategy 2 we took into account VerbNet roles. The first evaluation method is based on CoNLL2009 score which takes into account the head words only, and PropBank roles. The latter are much lower in number with respect to their corresponding VerbNet elements. Moreover, matching the head word only for a certain filler probably oversimplifies the matches. 


 The main lesson learned is that NLP evaluation settings may be inadequate when measuring the absolute performance of a semantic task as complex as SRL. Since the contrastive results show that the differences in method performance are consistent, even if at different accuracy levels, the accuracy seems to entirely depend on the ``resolution'' or sensibility of the setting. We recommend to define knowledge-graph-oriented benchmarks and scorers, and, in the particular case of SRL, to revisit the way role ontologies are designed. 

\section{Conclusions and Future Work}\label{conclusions}
In this paper we have addressed the problem of frame parsing jointly using Framester and Stanford CoreNLP in a novel implemented algorithm, TakeFive. In particular, we aimed at detecting verb frames and their labeled arguments (semantic roles and fillers). In order to assess the quality of our approach, we have carried out a comparative performance evaluation between TakeFive, and other SRL tools including SEMAFOR, FRED, Pikes, PathLSTM. TakeFive, the only one using a hybrid knowledge-based approach, is close to the best with the Wall Street Journal corpus from the Penn TreeBank. We have also observed that a simple ensemble of TakeFive, and the FRED machine reader, produces the best overall results. 

We have noticed that natural texts (even in the reasonably controlled production of the Wall Street Journal) contain many more linguistic phenomena than expected in existing manually developed resources such as VerbNet (for this reason, FRED uses a greedy algorithm for SRL instead of one that is closed under one specific resource). 

As ongoing and future work, we aim at designing new role ontologies that respond to best practices in knowledge graph design. We also want to use ensemble learning approaches by combining multiple methods, and feeding/controlling the ensemble pipeline by using existing linguistic resources for SRL, and heuristical methods. 



\section*{References}
\bibliographystyle{elsarticle-num}
\bibliography{references}

\end{document}